\documentclass[runningheads,a4paper]{llncs}

\usepackage{amssymb}
\usepackage{graphicx}
\usepackage{multirow}
\usepackage{booktabs}
\usepackage{xcolor}
\usepackage{url}
\usepackage[hidelinks]{hyperref} 
\usepackage{amssymb}
\usepackage{amsmath, xparse}

\urldef{\mailsa}\path|sossistefano7@gmail.com,|
\urldef{\mailsb}\path|gv@urubo.org,|
\urldef{\mailsc}\path|dzieba@wne.uw.edu.pl|    
\newcommand{\keywords}[1]{\par\addvspace\baselineskip
\noindent\keywordname\enspace\ignorespaces#1}

\begin{document}

\mainmatter  

\title{A Machine Learning Approach For Bitcoin Forecasting}

\titlerunning{Bitcoin Forecasting}

%
%
\author{Stefano Sossi-Rojas$^1$%
\and Gissel Velarde$^1$\and Damian Zieba$^2$}
\authorrunning{Sossi-Rojas, Velarde, Zieba}

\institute{$^1$Universidad Privada Boliviana, Computational Systems Engineering,\\
Cochabamba, Bolivia\\
$^2$University of Warsaw, Faculty of Economic Sciences, \\
Poland\\
\mailsa\\
\mailsb\\
\mailsc\\
}

%
%

\toctitle{Lecture Notes in Computer Science}
\tocauthor{Authors' Instructions}
\maketitle

\begin{abstract}
Bitcoin is one of the cryptocurrencies that is gaining more popularity in recent years.
Previous studies have shown that closing price alone is not enough to forecast stock market series. We introduce a new set of time series and demonstrate that a subset is necessary to improve directional accuracy based on a machine learning ensemble. In our experiments, we study which time series and machine learning algorithms deliver the best results. We found that the most relevant time series that contribute to improving directional accuracy are Open, High and Low, with the largest contribution of Low in combination with an ensemble of Gated Recurrent Unit network and a baseline forecast. The relevance of other Bitcoin-related features that are not price-related is negligible. The proposed method delivers similar performance to the state-of-the-art when observing directional accuracy. 

\keywords{Bitcoin, Forecasting, Time Series, Machine Learning}
\end{abstract}

\section{Introduction}

Bitcoin price forecasting has been the aim of many studies in the literature over the years. Nevertheless, unexpected price movements, smaller and larger bubbles, and different short- and long-term trends keep this task an ongoing topic for research. In one of the recent studies exploring this area of research \cite{chevallier2021possible}, authors verify the performance of different machine-learning algorithms and mention the current state of the knowledge on Bitcoin forecasting. One aspect is that referring to \cite{baur2019price} the price of Bitcoin is mainly driven by spot market rather than futures market. The other aspect is the division of approaches to forecasting into “Blockchain approach” and “Financial market’s approach”. The former is based on technical variables such as hash rates or mining difficulty, while the latter uses standard econometric variables such as stocks, bonds, or gold price. What can be added to that is that the other approach is to take variables related to investor sentiment such as Google trends data \cite{kristoufek2013bitcoin}, uncertainty indices (VIX, UCRY, see \cite{lucey2022cryptocurrency}), or the Fear and Greed index which has not been yet much explored in the literature.
In our study, we use the mix of features from those three approaches as well as the Open, High, Low and Close prices to predict the Bitcoin price using a novel approach. Based on that, we are able to verify which variables contribute to the forecast accuracy the most. In fact, the features with the largest contribution are only the price-related ones. The other variables, whether coming from “Blockchain approach”, “Financial markets approach” or “Sentiment approach” have a negligible impact in improving Bitcoin’s price performance.

We use a method that explores features and machine learning algorithms for Bitcoin's closing price prediction. On recent forecasting competitions, it was observed that machine learning models and hybrid approaches demonstrated superiority over alternative methods \cite{makridakis2022m5}.  Therefore, in this study we evaluated the following machine learning algorithms: Long Short Term Memory (LSTM), Bidirectional Long Term Memory (BiLSTM), Gated Recurrent Unit (GRU), Bidirectional Gated Recurrent Unit (BiGRU), and Light Gradient Boosting Machine (LightGBM). In addition we use ensembling. The results were evaluated observing the predicted and actual Bitcoin closing price measuring Root Mean Squared Error (RMSE), Mean Squared Error (MSE), Mean Absolute Error (MAE) and Directional Accuracy (DA). 

Previous studies show that stock market price is not enough for prediction when training deep learning models \cite{velarde2022open},  \cite{Velarde2022}. In this work, we demonstrate that indeed Bitcoin closing price is not sufficient for forecasting and additional features are necessary when using machine learning algorithms. Evaluation on return, shows that the method developed in this work presents one of the highest scores with a directional accuracy score equal to 0.7645 exceeding a baseline by 58.24 percent.

We compare our work to related studies for Bitcoin prediction. In \cite{altan2019digital}, LSTM was used in combination to the Empirical Wavelet Transform (EWT) decomposition technique. The authors used the Intrinsic Mode Function (IMF) to optimize and estimate outputs with Cuckoo Search (CS) \cite{altan2019digital}.
In \cite{cohen2020forecasting}, Linear Regression (LR) techniques and particle swarm optimization
were used to train and forecast data from beginning of 2012 to the end of March 2020. The Best setup for the model was obtained with 42 days plus 1 standard deviation \cite{cohen2020forecasting}.
In  \cite{wirawan2019short}, Autoregressive Integrated Moving Average (ARIMA) was used for data from 1 May 2013 to 7 June 2019. This model works best for short term predictions and can be used to predict Bitcoin for one to seven days ahead \cite{wirawan2019short}.
Finally, in \cite{li2022hybrid}, a BiLSTM with Low-Middel-High features (LMH-BiLSTM) was tested with  two primary steps: data decomposition and bidirectional deep learning. Results demonstrate that the proposed model outperforms other benchmark models and achieved high investment returns in the buy-and-hold strategy in a trading simulation \cite{li2022hybrid}. 

In this work, we tested the previously mentioned machine learning algorithms one by one, and also in ensemble. We feed the algorithms with a new set of 13 time series, see section \ref{s:data_collection}. In addition, we included 11 signals that come from Variational Mode Decomposition (MMD) as proposed in \cite{li2022hybrid}. However, in our experiments, data decomposition did not provide significant improvements.  
Next, we explain our proposed method. 

\section{Method}
\begin{figure}
\centering
\includegraphics[height=5.6cm]{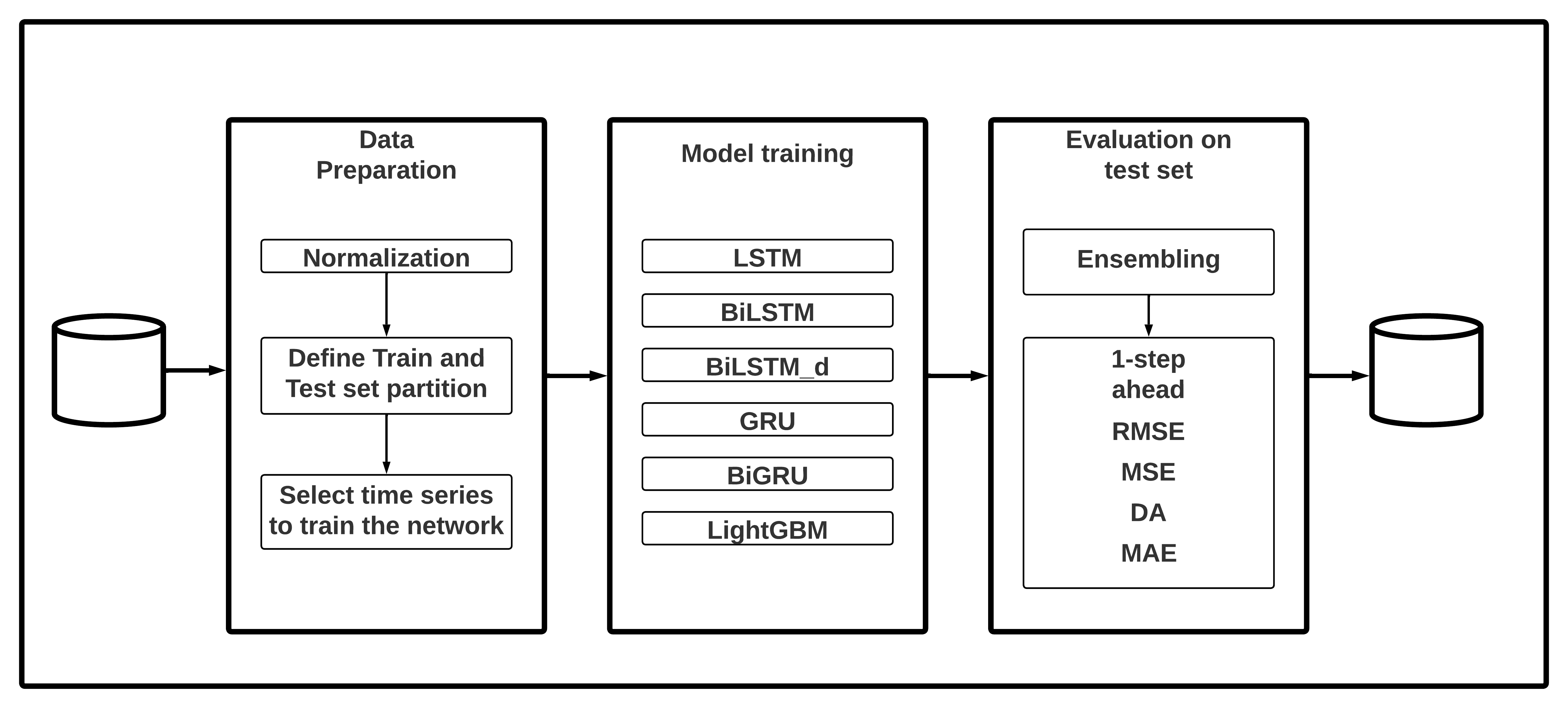}
\caption{Visual summary of the method.}
\label{fig:1}
\end{figure}
An overview of the method is presented in Fig. \ref{fig:1}. The input data is prepared in three steps. First, it is normalized between 0 and 1. Then, the train and test set partitions are created, where the train set is used for training several machine learning algorithms. A next phase considers algorithms' training. The details on hyper-parameter selection are described in section \ref{s:training}. Next, the evaluation phase is performed as rolling forecast for 1-step ahead prediction over the test set. 

\subsection{Data Collection} \label{s:data_collection}
We collected daily Bitcoin closing price from 7 October 2013 to 6 November 2022. We used a public API from the Kraken page \cite{Kraken}, the data collected are the values of: Close, Open, High, Low, Volume and Date. With the Nasdaq-Data-Link library for Python \cite{Nasdaq} the values were obtained: Transaction fee, Estimated Bitcoin USD Transaction Volume, Bitcoin USD Exchange Trade Volume, Bitcoin Hash Rate.  Bitcoin Google trends  were obtained with the pytrends library \cite{GoogleTrends}. The Gold to USD Exchange Rate was obtained from the Investing.com page. The Fear and Greed Index was obtained from the Kaggle page  \cite{Fear_Araujo}. The moving average of the Closing value was added, taking the last 30 days. Table \ref{Table1} presents all features used in this study. A similar set of features has been used in \cite{li2022hybrid}: Bitcoin Price, Bitcoin Transaction fees as Bitcoin miner’s revenue divided by transactions, USD trade volume from the top Bitcoin exchanges, Bitcoin transaction volume, USD exchange or trade volume from the top Bitcoin exchanges, Gold exchange rate to US dollar, Hash rate, and Google Trends of Bitcoin.

\begin{table}[]
\centering
\caption{Series, Short Name, Description, and Count. Originally some signals had more samples than others, therefore, for those dates where no information was recorded for all signals, those dates were removed to obtain the same number of samples per signal down sampling to 3812 in the range 19 November 2013 to 4 November 2022.  }\label{Table1}
\begin{tabular}{p{0.1\textwidth}p{0.2\textwidth}p{0.6\textwidth}p{0.1\textwidth}}
Series      & Features        & Description                                                                                      & Count \\  \toprule
Series 1  & Close           & Daily Bitcoin close price                                                                         & 4379  \\
Series 2  & Low             & Daily Bitcoin low price                                                                           & 4379  \\
Series 3  & High            & Daily Bitcoin high price                                                                          & 4379  \\
Series 4  & Open            & Daily Bitcoin open price                                                                          & 4379  \\
Series 5  & Trans\_Volume   & Bitcoin transaction volume in dollars                                                             & 3812  \\
Series 6  & Volume          & Daily quantity of Bitcoins bought or sold                                                         & 4379  \\
Series 7  & Hash\_Rate      & Number of giga hashes Bitcoin network performed                                                   & 3812  \\
Series 8  & Trans\_Fees     & Bitcoin miner's revenue divided by transactions                                                   & 3812  \\
Series 9  & XAU\_USD        & Gold (XAU) Exchange rate to US dollar (USD)                                                       & 3812  \\
Series 10 & Trade\_Volume   & Bitcoin trade volume in dollars                                                                   & 3812  \\
Series 11 & Google\_Trend   & Bitcoin's Google Trend                                                                           & 3812  \\
Series 12 & Fear\_Greed     & Fear \& Greed Index. It is a way to gauge stock market movements and whether stocks are fairly priced & 3812  \\
Series 13 & Moving\_Avg\_30 & Moving average of Bitcoin's closing price for the last 30 days.                                     & 3783 \\   \bottomrule
\end{tabular}

\end{table}

\subsection{Variational Mode Decomposition (VMD)}
\begin{figure}
\centering
\includegraphics[height=11cm]{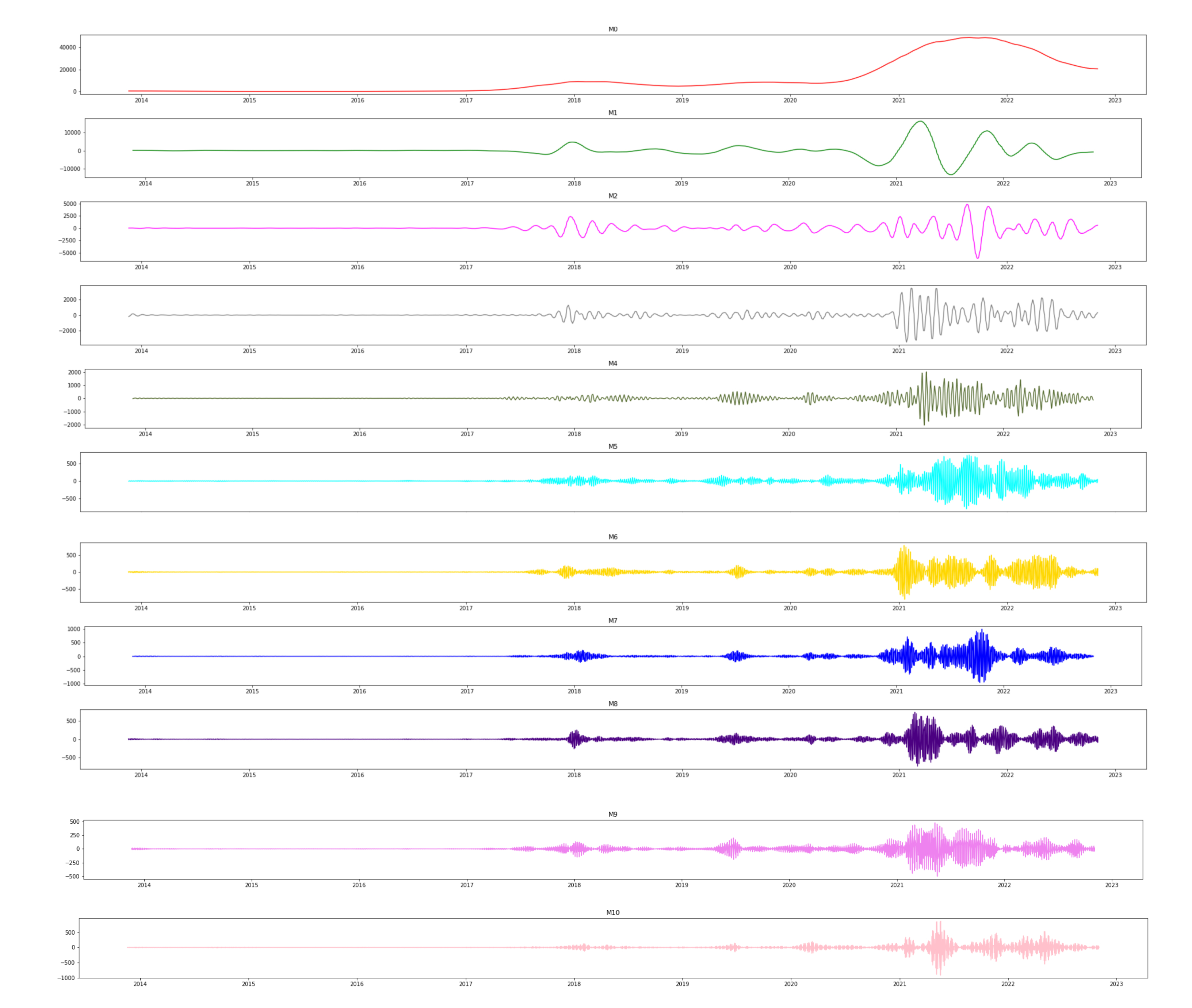}
\caption{Variational Mode Decomposition (VMD) decomposition of Bitcoin's closing price from 7 October 2013 to 6 November 2022.}
\label{fig:2}
\end{figure}

Bitcoin' closing price was decomposed using VMD method as proposed in \cite{li2022hybrid}. Each decomposed mode is labeled M0 through M10, respectively, where M0 has the lowest frequency and M10 has the highest. We can observe the graph of the decomposition in Fig.  \ref{fig:2}.  

Variational Mode Decomposition (VMD) is a completely non-recursive signal decomposition technique proposed by \cite{Konstantine}. VMD is a problem of
variational optimization that aims to minimize the total bandwidth of each mode.
This work used the vmdpy python library \cite{Vinicius}, with parameters by default using Bitcoin's Close price as input and a bandwith of 5000.

\begin{figure}
\centering
\includegraphics[height=6cm]{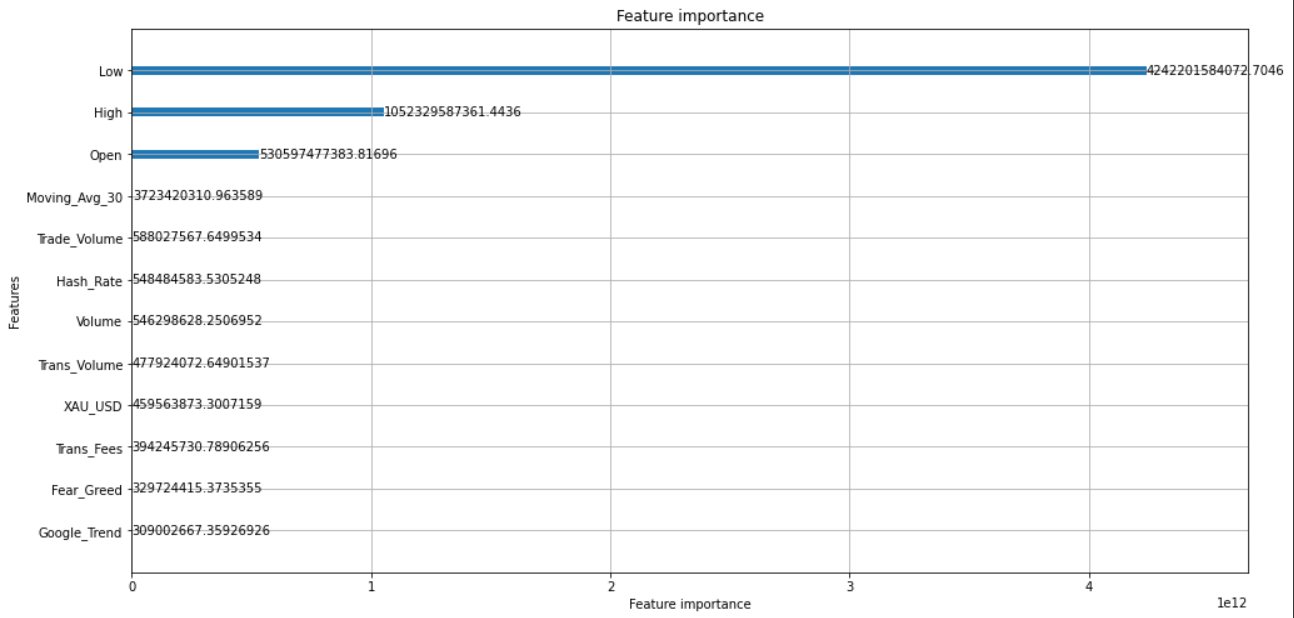}
\caption{Feature importance found by LightGBM. Most important features are Low, High and Open.} 
\label{fig:3}
\end{figure}

\subsection{Models' Training} \label{s:training}
We tested five deep learning arquitectures and one tree boosting method. All deep learning networks present an input layer of 90 units. A set value of 500 epochs and a batch size of 64 without early stopping.  

\begin{itemize}
\item \textbf{Long-Short Term Memory (LSTM)}: The network trains with 5 layers an input layer with the activation function 
seen in [6], bias initializer Gloroth Uniform, kernel regulator l1,l2, kernel constraint unit norm and the time\_major activated. Follow a dense layer of 90 units and linear activation. An output layer and another dense layer. The model uses the Adam optimizer with a learning rate of 0.002.
\item \textbf{Gated Recurrent Unit (GRU)}: The network trains with 5 layers, an input layer followed by a dropout layer set to 0.3, an output layer followed by another dropout layer, and a dense layer of one unit. The model uses the Adam optimizer with a learning rate of 0.0001. A similar network has been used in \cite{Niousha_Rasifaghihi}.
\item \textbf{Bidirectional Long-Short Term Memory (BiLSTM)}: The network consists of an input layer with the \textit{tanh} activation function, followed by a backward learning layer and a dense layer. The model uses the Adam optimizer with a learning rate of 0.01. A similar network has been used in \cite{li2022hybrid}
.
\item \textbf{Bidirectional Long-Short Term Memory with dropout (BiLSTM\_d)}: The network consists of an input layer with \textit{tanh} activation function, followed by a dropout layer, followed by a backward learning layer, followed by a dropout layer, and a dense layer. The model uses  Adam optimizer with a learning rate of 0.01 and the dropout set to 0.3.
\item \textbf{Bidirectional Gated Recurrent Unit (BiGRU)}: The network trains with 5 layers, an input bidirectional layer followed by a dropout layer set to 0.3, an output bidirectional layer followed by another dropout layer and a dense layer. The model uses Adam optimizer with a learning rate of 0.0001.

\item \textbf{Light Gradient Boosting Machine (LighGBM)}: Presents an early stopping round set to 50, verbose evaluation set to 30 with 3600 number of boost rounds.  The model trains with a gradient booting decision tree, objective set to \textit{tweedie} with a variance power of 1.1, uses an RMSE metric with n-jobs set to -1. Beside, 42 seeds with a learning rate of 0.2, the bagging fraction is set to 0.85 and the bagging frequency is set to 7, colsample by tree and colsample by node are set to 0.85 with a min data per leaf of 30, number of leaves is 200 with lambda \textit{l1} and \textit{l2} set to 0.5. A similar network has been used in \cite{makridakis2022m5}. 

\end{itemize}

\subsection{Evaluation}
Predicted results are normalized between 0 and 1 before evaluation measurements are made. 

We measure Mean Squared Error (MSE), Root Mean Squared Error (RMSE), Mean Absolute Error (MAE), and Directional Accuracy (DA) between the predicted and actual closing price as described in \cite{wang2012stock}, such that $n$ is the number of samples, $y_t$ and $x_t$ are predicted and actual closing price at time $t$:

\begin{equation}  \label{e:1}
MSE = n^{-1}\sum_{t=1}^{n}(x_t-y_t)^2.
\end{equation}

\begin{equation}
RMSE = \sqrt{MSE}.
\end{equation}

\begin{equation}
MAE=\frac{ \sum_{t=1}^{n}|y_t-x_t|}{n}.
\end{equation}

\begin{equation}
DA = \frac{100}{n}\sum_{t=1}^{n}d_t,
\end{equation}
where:
\begin{equation*}
d_t=\left\{\begin{matrix}
1 & (x_{t}-x_{t-1})(y_{t}-y_{t-1})\geq 0\\ 
0 & otherwise.
\end{matrix}\right.
\end{equation*}

\subsection{Return}
 The return is a financial measure used to assess the efficiency of an asset investment. It is an growth indicator of the value of an investment during a certain period of time. Return On Investment (ROI) is one of the main financial measures used both in the traditional stock market and in the world of cryptocurrencies \cite{Phemex}. The formula can be expressed in terms of the Final Value of Inversion (FVI) and the Initial Value of Inversion (IVI):

\begin{equation} \label{eq:return}
ROI = (\frac{FVI - IVI}{IVI}) 100\%.
\end{equation}

\section{Experiments}
\subsection{Data preparation}
For the LSTM, BiLSTM, GRU and BiGRU models, the data was scaled between 0 and 1 before training, except LightGBM model for which data scaling was done for evaluation only. 
Next, the data was divided into 25-day windows, converting the data table into 3-D lists (arrays), where the first dimension corresponds to batch size, the second to number of time-steps, and finally, the third dimension to the number of units of one input sequence \cite{ShivaVerma}.
Next to each list the expected future value was saved. This last value is taken from the closing values of the previous day. For the LightGBM model, the complete data Table \ref{Table1} was used without modifications.
The data partitioning was done as follows: train set from 7 October 2013 to 8 August 2022, test set from 9 August 2022 to 6 November 2022. Where 90 days was used for testing, and rest of data for training. 

\subsection{Experiments and Results}
Experiments were conducted to see how different time series influence Bitcoin closing price prediction and how different models perform. The experiments were carried out analyzing the results with different factors.
Tables \ref{Table2} to \ref{Table11} show the results obtained in the different experiments. RMSE, MSE, MAE close to zero and DA close to one are preferred. The best results are highlighted. The values are compared with values obtained with a Baseline prediction. Baseline prediction means that the predicted value is the last observed value. Series importance for the prediction was obtained according to the LightGBM model to later classify them and continue with the experiments as can be seen in Fig. \ref{fig:3}.
We performed the following experiments:

\begin{itemize}
\item \textbf{Experiment 1.} The first experiment sends a subset of all tested time series, as we assumed these features represent price action over a set period of time, and in combination could be used to predict price movements. The time series used are: Close, Open, High, Low and Volume of Bitcoin.  These are obtained from the Kraken API \cite{Kraken}.  See Table \ref{Table2} for results of prediction. Table \ref{Table2} shows that BiLSTM has the best DA = 0.5056 but BiGRU presents the lowest MAE = 0.0467.

\item \textbf{Experiment 2.} In the second experiment we test all time series presented in Table \ref{Table1}: Close, Open, High, Low, Volume, Transaction Fee, Estimated Bitcoin USD Transaction Volume, Bitcoin USD Exchange Trade Volume, Rate of Bitcoin Hash, Bitcoin Google Trends, and Gold to USD Exchange Rate, Fear and Greed Index and the Moving Average of the Closing value. The main idea was to enrich the input data to help improve prediction. The results can be seen in Table \ref{Table3}. LightGBM has the best performance and there is an improvement in comparison with Experiment 1 were less input data was used.

\item \textbf{Experiments 3.} For the following experiments, series importance for prediction was analyzed according to LightGBM model. The order of importance can be seen in Figure \ref{fig:3}. The experiments were performed giving different combinations of the four most important series. Open, High, Low and Close value. These time series have a great correlation with the closing price of bitcoin and influence more in the behavior that it will have. For this experiment  Open, High and Low values were used. Table \ref{Table4} shows that BiLSTM has the best DA = 0.4832 but GRU has best MAE = 0.0496. Results show a decrease in performance.

\item \textbf{Experiment 4.} For this experiment High and Low values were used. Table \ref{Table5} shows that BiLSTM has the best performance with DA = 0.5169. There is an improvement in comparison with Experiment 1 but does not reach the performance of Experiment 2.

\item \textbf{Experiment 5.} For this experiment Low values were used. Table \ref{Table6} shows that LSTM has the best performance with DA = 0.5169. Results show a similar performance in comparison with Experiment 4.

\item \textbf{Experiment 6.} For this experiment Open and Low values were used. Table \ref{Table7} shows that LSTM has the best DA = 0.5393 and BiLSTM delivers the lowest errors. Results show an improvement in performance compared with previous experiments but no greater than that in Experiment 2.

\item \textbf{Experiment 7.} Since Experiment 2 had the best performance so far, we tested all time series used in Experiment 2 and added 11 VMD modes see their impact for prediction. That is, the input data includes the values of Close, Open, High, Low, Volume, Transaction Fee, Estimated Bitcoin USD Transaction Volume, Bitcoin USD Exchange Trade Volume, Rate of Bitcoin Hash, Bitcoin Google Trends, and Gold to USD Exchange Rate, Fear and Greed Index and the Moving Average of the Closing value and 11 VMD modes. Results are presented in Table \ref{Table8}. This time, BiGRU has the best DA = 0.5730 but LSTM has the lowest errors. Results show a big improvement in performance compared with previous experiments.

\item \textbf{Experiment 8.} This experiment consists of sending as input data the values all 11 VMD modes, that is only the 11 VMD modes were added to see the impact these had in the prediction. See Table \ref{Table9}. BiGRU has the best DA = 0.5618, BiLSTM the best MAE and LSTM the best MSE and RMSE. Results show an improvement in performance compared with previous experiments but no greater than that of Experiment 7.

\end{itemize}
We can observe in all the experiments a noticeable improvement of the BiLSTM model when it does not present the dropout layer.


\begin{table}[]
\centering
\caption{Prediction Measured Values with Bitcoin Close, Open, High, Low and Volume values.}\label{Table2}
\begin{tabular}{lllll} \toprule
\multirow{2}{*}{\textbf{Network}} & \multicolumn{4}{l}{\textbf{Measured value}}               \\
                                  & \textbf{MAE} & \textbf{MSE} & \textbf{RMSE} & \textbf{DA} \\ \midrule
GRU                               & 0.0541       & 0.0111       & 0.1053        & 0.4494      \\
BiGRU                             & \textbf{0.0467}       & 0.0107       & 0.1032        & 0.4494      \\
LSTM                              & 0.0660       & 0.0125       & 0.1120        & 0.4719      \\
BiLSTM                            & 0.0954       & 0.0163       & 0.1277        & \textbf{0.5056}      \\
BiLSTM\_d                         & 0.1610       & 0.0667       & 0.2583        & 0.4832      \\
LightGBM                          & 0.0513       & \textbf{0.0106}       & \textbf{0.1028}        & 0.4157  \\   \bottomrule
\end{tabular}
\end{table}

\begin{table}[]
\centering
\caption{Prediction measured values with all factors.}\label{Table3}
\begin{tabular}{lllll} \toprule
\multirow{2}{*}{\textbf{Network}} & \multicolumn{4}{l}{\textbf{Measured value}}               \\
                                  & \textbf{MAE} & \textbf{MSE} & \textbf{RMSE} & \textbf{DA} \\
GRU                               & 0.1426       & 0.0496       & 0.2228        & 0.3371      \\
BiGRU                             & 0.1501       & 0.0660       & 0.2569        & 0.3596      \\
LSTM                              & 0.1731       & 0.0666       & 0.2580        & 0.3708      \\
BiLSTM                            & 0.2019       & 0.0864       & 0.2940        & 0.4382      \\
BiLSTM\_d                         & 0.2685       & 0.1109       & 0.3330        & 0.3371      \\
LightGBM                          & \textbf{0.0566}       & \textbf{0.0154}       & \textbf{0.1240}        & \textbf{0.5618}     \\   \bottomrule
\end{tabular}
\end{table}

\begin{table}[]
\centering
\caption{Prediction Measured Values with Bitcoin Open, High and Low Values.}\label{Table4}
\begin{tabular}{lllll} \toprule
\multirow{2}{*}{\textbf{Network}} & \multicolumn{4}{l}{\textbf{Measured value}}               \\
                                  & \textbf{MAE} & \textbf{MSE} & \textbf{RMSE} & \textbf{DA} \\
GRU                               & \textbf{0.0496}       & \textbf{0.0103}       & \textbf{0.1014}        & 0.4494      \\
BiGRU                             & 0.0508       & 0.0110       & 0.1050        & 0.4719      \\
LSTM                              & 0.1029       & 0.0352       & 0.1875        & 0.4382      \\
BiLSTM                            & 0.0812       & 0.0178       & 0.1334        & \textbf{0.4832}      \\
BiLSTM\_d                         & 0.1317       & 0.0571       & 0.2390        & 0.4607      \\
LightGBM                          & 0.0864       & 0.0157       & 0.1254        & 0.3483     \\   \bottomrule
\end{tabular}
\end{table}

\begin{table}[]
\centering
\caption{Prediction Measured Values with High and Low values of Bitcoin.}\label{Table5}

\begin{tabular}{lllll} \toprule
\multirow{2}{*}{\textbf{Network}} & \multicolumn{4}{l}{\textbf{Measured value}}               \\
                                  & \textbf{MAE} & \textbf{MSE} & \textbf{RMSE} & \textbf{DA} \\
GRU                               & 0.0572       & 0.0112       & 0.1059        & 0.4157      \\
BiGRU                             & 0.0487       & 0.0112       & 0.1060        & 0.4607      \\
LSTM                              & 0.0508       & 0.0117       & 0.1083        & 0.4832      \\
BiLSTM                            & \textbf{0.0402}       & \textbf{0.0092}       & \textbf{0.0961}        & \textbf{0.5169}      \\
BiLSTM\_d                         & 0.1709       & 0.0599       & 0.2447        & 0.4832      \\
LightGBM                          & 0.0796       & 0.0217       & 0.1472        & 0.2472     \\   \bottomrule
\end{tabular}
\end{table}

\begin{table}[]
\centering
\caption{ Prediction Measured Values with Low value of Bitcoin.}\label{Table6}

\begin{tabular}{lllll} \toprule
\multirow{2}{*}{\textbf{Network}} & \multicolumn{4}{l}{\textbf{Measured value}}               \\
                                  & \textbf{MAE} & \textbf{MSE} & \textbf{RMSE} & \textbf{DA} \\
GRU                               & 0.0558       & 0.0116       & 0.1076        & 0.4719      \\
BiGRU                             & 0.5616       & 0.0117       & 0.1080        & 0.4607      \\
LSTM                              & \textbf{0.0378}       & \textbf{0.0098}       & \textbf{0.0988}        & \textbf{0.5169}      \\
BiLSTM                            & 0.0460       & 0.0104       & 0.1019        & 0.4719      \\
BiLSTM\_d                         & 0.1155       & 0.0523       & 0.2286        & 0.4944      \\
LightGBM                          & 0.1213       & 0.5991       & 0.2448        & 0.1348     \\   \bottomrule
\end{tabular}
\end{table}

\begin{table}[]
\centering
\caption{Prediction Measured Values with Bitcoin Open and Low Values.}\label{Table7}

\begin{tabular}{lllll} \toprule
\multirow{2}{*}{\textbf{Network}} & \multicolumn{4}{l}{\textbf{Measured value}}               \\
                                  & \textbf{MAE} & \textbf{MSE} & \textbf{RMSE} & \textbf{DA} \\
GRU                               & 0.5125       & 0.0102       & 0.1011        & 0.4607      \\
BiGRU                             & 0.0518       & 0.0104       & 0.1022        & 0.4494      \\
LSTM                              & 0.0613       & 0.0151       & 0.1228        & \textbf{0.5393}      \\
BiLSTM                            & \textbf{0.0344}       & \textbf{0.0092}       & \textbf{0.0958}        & 0.4719      \\
BiLSTM\_d                         & 0.1221       & 0.0623       & 0.2496        & 0.5056      \\
LightGBM                          & 0.0801       & 0.0227       & 0.1505        & 0.2809     \\   \bottomrule
\end{tabular}
\end{table}

\begin{table}[]
\centering
\caption{Prediction Measured Values with all factors plus VMD modes}\label{Table8}

\begin{tabular}{lllll} \toprule
\multirow{2}{*}{\textbf{Network}} & \multicolumn{4}{l}{\textbf{Measured value}}               \\
                                  & \textbf{MAE} & \textbf{MSE} & \textbf{RMSE} & \textbf{DA} \\
GRU                               & 0.1002       & 0.0245       & 0.1565        & 0.5506      \\
BiGRU                             & 0.0751       & 0.0230       & 0.1517        & \textbf{0.5730}      \\
LSTM                              & \textbf{0.0705}       & \textbf{0.0206}       & \textbf{0.1423}        & 0.5169      \\
BiLSTM                            & 0.0963       & 0.0222       & 0.1489        & 0.4832      \\
BiLSTM\_d                         & 0.1081       & 0.0322       & 0.1793        & 0.4832      \\
LightGBM                          & 0.0740       & 0.0222       & 0.1490        & 0.4719     \\   \bottomrule
\end{tabular}
\end{table}

\begin{table}[]
\centering
\caption{Prediction Measured Values with 11 VMD modes.}\label{Table9}

\begin{tabular}{lllll} \toprule
\multirow{2}{*}{\textbf{Network}} & \multicolumn{4}{l}{\textbf{Measured value}}               \\
                                  & \textbf{MAE} & \textbf{MSE} & \textbf{RMSE} & \textbf{DA} \\
GRU                               & 0.0806       & 0.0179       & 0.1338        & 0.5169      \\
BiGRU                             & 0.0762       & 0.0176       & 0.1327        & \textbf{0.5618}      \\
LSTM                              & 0.0591       & \textbf{0.0167}       & \textbf{0.1290}        & 0.4607      \\
BiLSTM                            & \textbf{0.0521}       & 0.0176       & 0.1329       & 0.4607      \\
BiLSTM\_d                         & 0.0837       & 0.0209       & 0.1446        & 0.5169      \\
LightGBM                          & 0.1299       & 0.0486       & 0.2204        & 0.4494     \\   \bottomrule
\end{tabular}
\end{table}

\subsection{Ensembling and Return Performance}
Since we are predicting one day in the future (rolling forecast), the value of IVI seen in equation \ref{eq:return}, is the same value of Baseline and FVI is the prediction of the models. Ensembling was obtained as the the simple arithmetic average, using equation \ref{eq:return}, combining each model with the baseline. We tested each model at a time. The results show that the best combination was GRU and the baseline training the model using the Open, High and Low factors.
\begin{itemize}
\item \textbf{Experiment 9.} In this experiment we test ensembling. Return values were calculated over the prediction of all previous experiments. The best results were obtained using Open, High and Low values as input. See Table \ref{Table10} for results. GRU has the best performance with DA = 0.7645. Results show a big improvement in DA  compared to previous experiments.

\item \textbf{Experiment 10.} The last experiment was executed to have a close comparison with LMH-BiLTSM \cite{li2022hybrid}, and therefore, we used approximately the same date range as used in \cite{li2022hybrid}. Return values were calculated over the prediction period. See Table \ref{Table11}. This time, GRU has the best DA = 0.7865 but BiGRU deliver the lowest errors. The DA obtained here is comparable to that reported by LMH-BiLTSM \cite{li2022hybrid}. Indeed, in this experiment, we observe the highest DA prediction of all experiments we performed. However, the lowest errors are measured using BiLSTM network with Open and Low series: MAE = 0.0344, MSE = 0.0092, and RMSE = 0.0958, see Table \ref{Table7}.
\end{itemize}

Table \ref{Table12} shows a comparison between previous studies and the best performing model presented in this paper. The model with the best performance consists of a GRU trained with Open, High and Low values of Bitcoin. The GRU ensemble achieves 0.7865  $\pm$   0.2113  DA, matching the performance obtained by LMH-BiLSTM \cite{li2022hybrid} in a similar time range. Notice that for the period October 2013 to 6 November 2022, the DA is slightly lower but the standard deviation is also lower. 
Making a comparison of the GRU network (Table \ref{Table10}) and the Baseline we see an improvement of 58.14 percent in directional accuracy, being the model with the highest score. 

\begin{table}[]
\centering
\caption{Return Measured Values for prediction with Bitcoin Open, High and Low Values. Range between 7 October 2013 and 6 November 2022.}\label{Table10}

\begin{tabular}{lllll} \toprule
\multirow{2}{*}{\textbf{Network}} & \multicolumn{4}{l}{\textbf{Measured value}}               \\
                                  & \textbf{MAE} & \textbf{MSE} & \textbf{RMSE} & \textbf{DA} \\
Baseline                               & 0.0371       & 0.0091       & 0.0952       & 0.4831      \\
GRU                               & \textbf{0.0843}       & \textbf{0.0296}       & \textbf{0.1716}        & \textbf{0.7645}      \\
BiGRU                             & 0.2299       & 0.0735       & 0.2711        & 0.7191      \\
LSTM                              & 0.2629       & 0.1113       & 0.3337        & 0.5169      \\
BiLSTM                            & 0.1687       & 0.0553       & 0.2352        & 0.5169      \\
BiLSTM\_d                         & 0.3082       & 0.1072       & 0.3274        & 0.7303      \\
LightGBM                          & 0.1730       & 0.0488       & 0.2208        & 0.6547     \\   \bottomrule
\end{tabular}
\end{table}

\begin{table}[]
\centering
\caption{Return Measured Values for prediction with all factors plus VMD modes in 7 October 2013 to 1 January 2021 range approximately same range seen in \cite{li2022hybrid}. Baseline DA is 0.5281.}\label{Table11}

\begin{tabular}{lllll} \toprule
\multirow{2}{*}{\textbf{Network}} & \multicolumn{4}{l}{\textbf{Measured value}}               \\
                                  & \textbf{MAE} & \textbf{MSE} & \textbf{RMSE} & \textbf{DA} \\
Baseline                               & 0.0496       & 0.0138       & 0.1176        & 0.5281      \\
GRU                               & 0.1745       & 0.0548       & 0.2341        & \textbf{0.7865}      \\
BiGRU                             & \textbf{0.1137}       & \textbf{0.0400}       & \textbf{0.2001}        & 0.7303      \\
LSTM                              & 0.2039       & 0.0860       & 0.2933        & 0.5955      \\
BiLSTM                            & 0.2772       & 0.1044       & 0.3232        & 0.6067      \\
BiLSTM\_d                         & 0.3134       & 0.1332       & 0.3649        & 0.5056      \\
LightGBM                          & 0.2806       & 0.1078       & 0.3283        & 0.5730     \\   \bottomrule
\end{tabular}
\end{table}

\begin{table}[]
\centering
\caption{Performance Comparisons. The methods presented in this table were trained using different datasets in different date ranges. Therefore, this comparison is a relative. 
}\label{Table12}

\includegraphics[height=3.2cm]{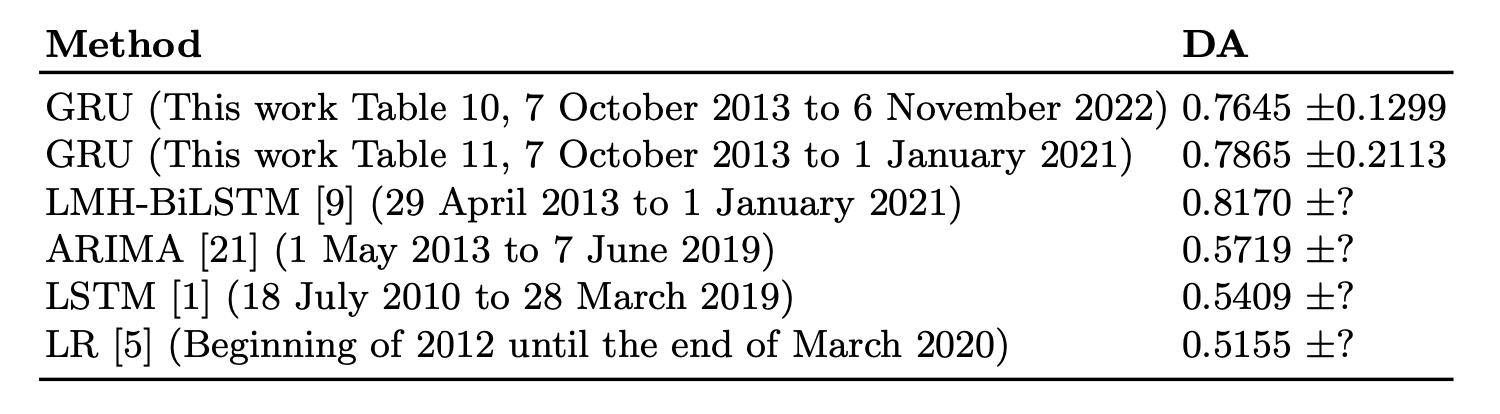}

%
\end{table}
\newpage

\section{Conclusion}
We confirm the hypothesis that Bitcoin is difficult to predict with the closing price alone, that is, the closing price does not contain enough information to predict Bitcoin, and a set of time series are necessary to improve prediction. We tested 13 series as shown in Table \ref{Table1}, plus 11 modes decomposed using Variational Mode Decomposition (VMD). In addition, we tested various machine learning algorithms, and found that a selected set of time series consisting of Open, High, Low values and an ensemble based on a GRU network combined to the value of return, or a baseline prediction, demonstrates a great improvement in the results of the experiments. Our method delivers a comparable DA when compared to the state-of-the-art, which in contrast uses a BiLSTM with Low-Middel-High features (LMH-BiLSTM) \cite{li2022hybrid}.

\bibliographystyle{splncs03}
\bibliography{References.bib}
%








\end{document}